\documentclass{article}
\usepackage{spconf,amsmath,graphicx}
\usepackage{hyperref}
\usepackage{url}
\usepackage{graphicx}
\usepackage{booktabs}
\usepackage{xspace}

\newcommand{\methodname}{RetCLIP\xspace}

\title{Open Vocabulary Panoptic Segmentation With Retrieval Augmentation}

\name{}
\address{}

\twoauthors
 {Nafis Sadeq\sthanks{Work done during internship at Samsung Semiconductor, Inc.}}
	{East West University\\
	nafis.sadeq@ewubd.edu}
 {Qingfeng Liu, Mostafa El-Khamy}
	{Samsung Semiconductor, Inc.\\
	\{qf.liu,mostafa.e\}@samsung.com}

\begin{document}
\maketitle
\begin{abstract}
Given an input image and set of class names, panoptic segmentation aims to label each pixel in an image with class labels and instance labels. In comparison, Open Vocabulary Panoptic Segmentation aims to facilitate the segmentation of arbitrary classes according to user input. The challenge is that a panoptic segmentation system trained on a particular dataset typically does not generalize well to unseen classes beyond the training data. In this work, we propose \methodname, a retrieval-augmented panoptic segmentation method that improves the performance of unseen classes. In particular, we construct a masked segment feature database using paired image-text data. At inference time, we use masked segment features from the input image as query keys to retrieve similar features and associated class labels from the database. Classification scores for the masked segment are assigned based on the similarity between query features and retrieved features. The retrieval-based classification scores are combined with CLIP-based scores to produce the final output. We incorporate our solution with a previous SOTA method (FC-CLIP). When trained on COCO, the proposed method demonstrates 30.9 PQ, 19.3 mAP, 44.0 mIoU on the ADE20k dataset, achieving +4.5 PQ, +2.5 mAP, +10.0 mIoU absolute improvement over the baseline.
\end{abstract}
\section{Introduction}
\label{sec:intro}
Panoptic segmentation is a computer vision task that combines semantic segmentation and instance segmentation~\cite{Kirillov_2019_CVPR}. Semantic segmentation labels every pixel in an image with a class category, such as "tree" or "car." Instance segmentation differentiates between individual objects of the same class (1st car, 2nd car). Panoptic segmentation unifies these tasks by labeling every pixel with a class label and identifying distinct objects within the same category with an instance label. This method is valuable in fields like autonomous driving and robotics, where detailed scene understanding is crucial. A key challenge for traditional panoptic segmentation is the need for highly granular pixel-level data annotation. Lack of data limits the number of possible classes for panoptic segmentation, making the system closed-vocabulary~\cite{ding2022open}.

Open vocabulary panoptic segmentation~\cite{ding2022open,xu2023masqclip,yu2024convolutions} is an advanced version of the traditional panoptic segmentation task that extends its capabilities to identify and label objects from a potentially unlimited set of classes. Unlike standard panoptic segmentation which relies on a fixed set of known classes, open vocabulary segmentation allows the system to recognize and categorize objects even if they haven't been specifically included in the training dataset.

Recent methods for open vocabulary segmentation~\cite{ding2022open,liang2023open,xu2023masqclip,yu2024convolutions} involves a two-stage framework. The first step is to generate a class-agnostic mask proposal and the second step is to leverage pre-trained vision language models (e.g., CLIP~\cite{radford2021learning}) to classify masked regions. In this approach, the input class descriptions are encoded with a CLIP text encoder, and the masked image region is encoded with a CLIP vision encoder. The masked region is classified based on the cosine similarity of masked image features and class-related text features. CLIP has shown the ability to improve open vocabulary performance because it is pre-trained to learn joint image-text feature representation from large-scale internet data. However, the performance of the CLIP vision encoder suffers from a limitation when we encode a masked image instead of a natural image. The domain shift between full image features and masked image features hurts open vocabulary segmentation performance~\cite{liang2023open}.

In this work, we address the bottleneck mentioned above in the context of open vocabulary panoptic segmentation. In order to mitigate the domain shift between the natural image feature and the masked image feature, we propose \methodname, a retrieval-augmented approach for panoptic segmentation. Specifically, we first use large-scale image-text pairs to construct a feature database with associated text labels for the masked regions. Then during inference time, the masked region feature extracted from the input image is used as a retrieval key to retrieve similar features and associated class labels from the database. The masked region is classified based on the similarity between the retrieval key and retrieval targets. Since both the retrieval key and retrieval target use a CLIP vision encoder on masked regions, the proposed approach does not suffer from the domain shift between the natural image feature and the masked image feature. We combine this retrieval-based classification module with the CLIP-based classification module to improve open vocabulary panoptic segmentation performance. Our contributions are as follows:
\begin{itemize}
\item We proposed \methodname, a retrieval-augmented panoptic segmentation approach that tackles the domain shift between the natural image feature and masked image feature with respect to the CLIP vision encoder. The proposed approach can incorporate new classes in the panoptic segmentation system simply by updating the feature database in a fully training-free manner. The feature database can be constructed from paired image-text data which is widely available for thousands of classes.
\item We demonstrate that the proposed system can improve open vocabulary panoptic segmentation performance in both training-free setup (+5.2 PQ) and cross-dataset fine-tuning setup (+ 4.5 PQ, COCO$\rightarrow$ADE20k).
\end{itemize}

\section{Methodology}

\subsection{Feature Database Construction} 
\begin{figure}
\centering
\includegraphics[width=\linewidth]{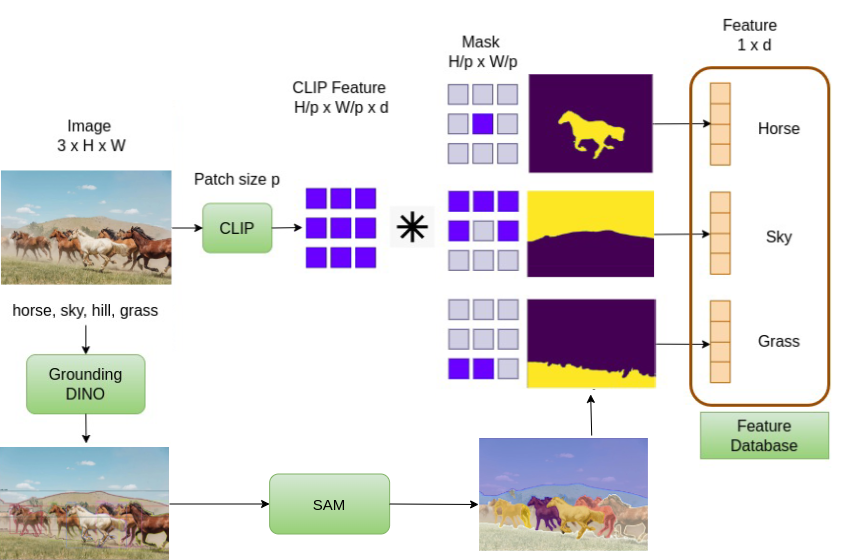}
\caption{Overview of feature database construction}
\label{fig:database}
\end{figure}

The objective of the feature database construction is to take a paired image-text dataset as input and convert it into a database of masked segment features and associated class labels. The database construction has four steps, namely object detection, mask generation, dense feature generation, and mask pooling. The overview of the process is shown in Figure~\ref{fig:database}.

\textbf{Object Detection: } In this step, an image and class labels present in the image are fed to an open vocabulary object detection method. The output is a bounding box associated with each object instance present in the image. We use the open vocabulary object detection method Grounding DINO~\cite{liu2024groundingdinomarryingdino}.

\textbf{Mask Generation: } In this step, the input image and associated bounding box prompts are fed to Segment Anything Model (SAM)~\cite{kirillov2023segment} for mask generation. Even though SAM can generate masks without class-aware bounding boxes, the resulting masks often break up a single class (e.g. car) into multiple masks (e.g. wheel, car body, window). The class-aware masks generated in the previous step ensure that the SAM can generate high-quality masks for each class present in the image.

\textbf{Dense Feature Generation: } We use CLIP to extract image-level dense features. Let's assume that the input image has shape $3\times H\times W$, the patch size of CLIP is $p$, and the dimension of the dense feature is $d$. The shape of the output dense feature is $\frac{H}{p} \times \frac{W}{p} \times d$.

\textbf{Mask Pooling: } Mask pooling operation involves taking dense features associated with the whole image and generating mask-specific dense features based on generated masks in the second step. This way we don't have to encode each masked segment using CLIP separately which can be computationally expensive~\cite{yu2024convolutions}. The mask pooling operation generates a $d$ dimensional feature vector for each masked segment. These features and associated class labels are added to the feature database.

\subsection{Panoptic Segmentation Method (\methodname)} 
\begin{figure*}
\centering
\includegraphics[width=0.8\linewidth]{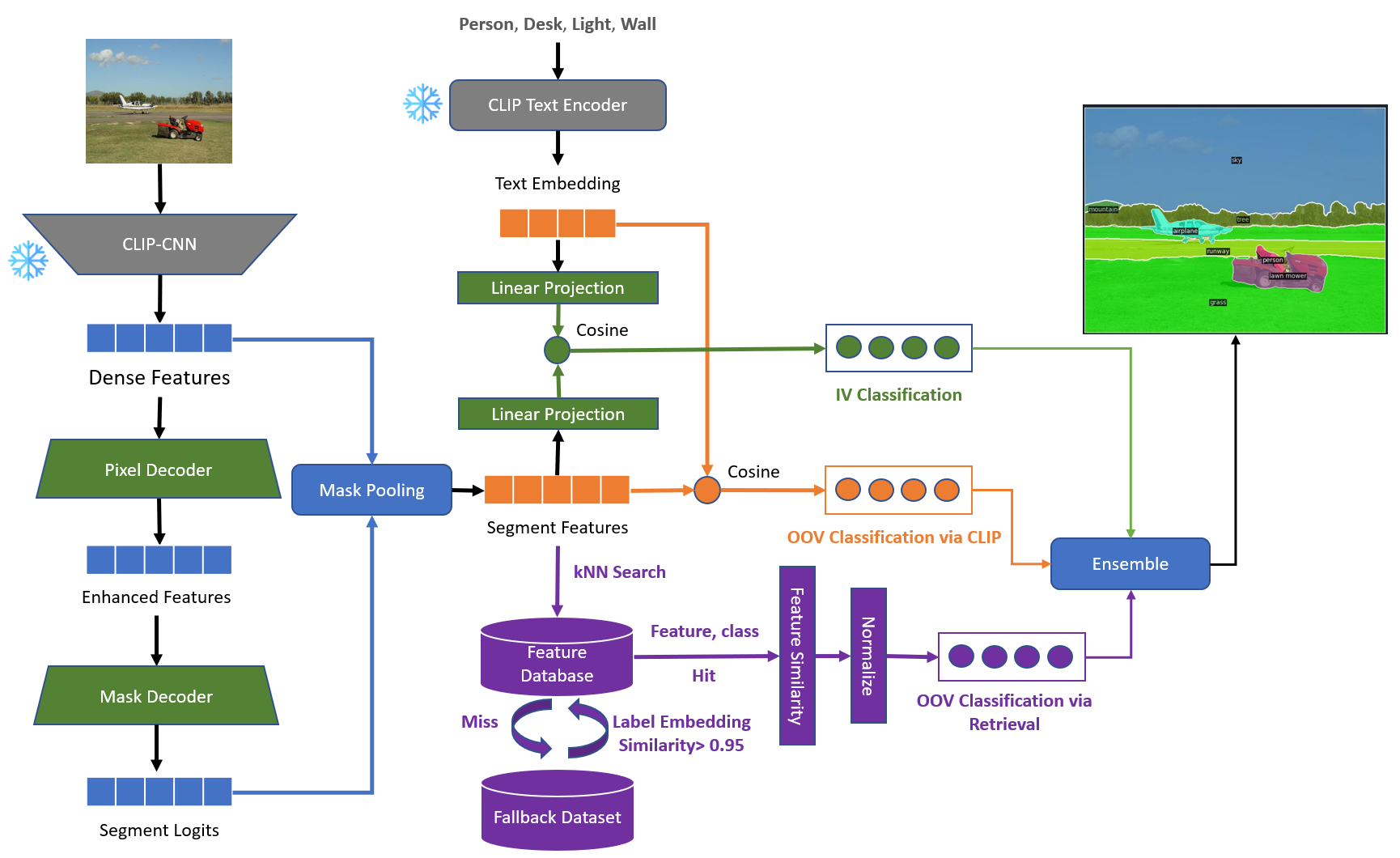}
\caption{Overview of \methodname. All components of the retrieval pipeline are shown in purple. The trainable components for cross-dataset fine-tuning are shown in green.}
\label{fig:cross_dataset}
\vspace{-1em}
\end{figure*}

In the cross-dataset setting of open vocabulary panoptic segmentation, the system is fine-tuned on one dataset (e.g. COCO) and evaluated on another dataset (ADE20k) with some unseen classes. Our cross-dataset method is based on FC-CLIP~\cite{yu2024convolutions} where a mask proposal generator and mask decoder are fine-tuned on COCO~\cite{lin2015microsoftcococommonobjects}. The overview of the system is shown in Figure~\ref{fig:cross_dataset}.

\textbf{Shared Backbone: } We use a frozen CNN-based CLIP backbone. The backbone is shared between the mask generation and segment classification. Yu et al. (2024)~\cite{yu2024convolutions} have demonstrated that the CNN-based CLIP backbone is a more robust variation in image resolution. We use the ConvNeXt-Large variant of CLIP backbones from OpenCLIP~\cite{Cherti_2023} in our best-performing system. The CLIP backbone converts the input image to patch-specific dense features which is used for mask generation and segment classification. 

\textbf{Mask Proposal Generation: } The mask proposal generator is based on Mask2former~\cite{cheng2022maskedattentionmasktransformeruniversal}. A pixel decoder is used for enhancing dense features from the CLIP backbone. The enhanced features and class-related queries are fed to a series of mask decoders. The mask decoders are equipped with self-attention, masked cross-attention, and a feed-forward network. Finally, the segmentation logits are produced via matrix multiplication between class queries and transformed pixel features. 

In the training-free variant of the system, we use Grounding DINO~\cite{liu2024groundingdinomarryingdino} to detect bounding boxes associated with each class. All bounding boxes detected with a minimum confidence threshold are retained. The bounding boxes are passed to SAM for generating class-aware masks. The outputs of SAM are used as class-agnostic mask proposals. All potential classes for panoptic segmentation are passed to the object detection method and confidence-based filtering is performed to prune absent classes.

\textbf{In Vocabulary Classification: } The in-vocabulary classification path is shown in green in Figure~\ref{fig:cross_dataset}. The dense features are computed from the input image feature and mask proposals using mask pooling. Dense features for masked regions and class name embeddings are projected to the same embedding space using linear projection. The linear projection parameters for in-vocabulary classifiers are fine-tuned on COCO. The classification scores are obtained based on cosine similarity between class embeddings and masked segment features.

\textbf{Out-of-vocabulary Classification Via Retrieval: } The retrieval-based classification path is shown in violet in Figure~\ref{fig:cross_dataset}. The retrieval module uses masked segment features as retrieval keys to perform approximate nearest neighbor search in the feature database. The output is a set of distance scores between the retrieval key and retrieval targets and associated class labels. The distance scores are normalized using min-max normalization and subtracted from one. This step produces retrieval-based classification scores. In case any of the user-provided class names are missing in the feature database, we retrieve image samples for those input classes from a secondary fallback image dataset. We use ADE20k to construct our primary feature database and the Google Open Image dataset as a fallback. The label matching between datasets is performed with CLIP text embedding of class names with similarity score $> 0.95$.

\textbf{Out-of-vocabulary Classification Via CLIP: } We have a CLIP-based classifier to handle out-of-vocabulary cases in addition to retrieval-based classification. This is helpful in case the feature database does not have any features associated with a user-provided class. The classification is performed using cosine similarity between segment features and class name embeddings. Unlike in-vocabulary classifiers, the features do not go through fine-tuned linear projection layers.

\textbf{Ensemble: } Let's assume $C$ is the set of classes for prediction and $C_{train}$ is the set of classes in the fine-tuning dataset. Let $s^i_{clip}, s^i_{ret}, s^i_{iv}$ be classification scores for class $i$ using CLIP, retrieval and in-vocabulary classifier. The scores from the three classification pipelines are combined as follows, where $\alpha, \beta, \gamma$ are hyper-parameters.
\begin{align*}
s^i_{oov} &= s^i_{ret}\times \gamma + s^i_{clip}\times (1 - \gamma) \\
s^i &= s^i_{oov}\times \alpha + s^i_{iv}\times (1 - \alpha) \texttt{ if}\ i \in C_{train}\\
s^i &= s^i_{oov}\times \beta + s^i_{iv}\times (1 - \beta) \texttt{ if}\ i \notin C_{train}\\
\end{align*}

\section{Evaluation}
\begin{table*}
\centering
\begin{tabular}{@{}llllllll@{}}
\toprule
\multicolumn{2}{l}{Method} & Backbone & Database & PQ    & mAP   & mIoU  \\ \midrule
\multicolumn{2}{l}{FC-CLIP} &  CLIP-RN50x64   & N/A   & 0.228 & 0.136 & 0.284  \\
\multicolumn{2}{l}{\methodname}   &  CLIP-RN50x64  & Google Open Image  & \underline{0.237} & \underline{0.141} & \underline{0.324} \\
\multicolumn{2}{l}{\methodname}  &  CLIP-RN50x64   & ADE20k   & \underline{\textbf{0.271}} & \underline{\textbf{0.151}} & \underline{\textbf{0.369}}  \\ \midrule
\multicolumn{2}{l}{FC-CLIP} &  CLIP-ConvNeXt-large   & N/A   & 0.264 & 0.168 & 0.340  \\
\multicolumn{2}{l}{\methodname}   &  CLIP-ConvNeXt-large  & Google Open Image  & \underline{0.283} & \underline{0.177} & \underline{0.383} \\
\multicolumn{2}{l}{\methodname}  &  CLIP-ConvNeXt-large   & ADE20k   & \underline{\textbf{0.309}} & \underline{\textbf{0.193}} & \underline{\textbf{0.440}}  \\
\bottomrule
\end{tabular}
\caption{Open vocabulary performance on ADE20k, after fine-tuning on COCO.}
\label{tab:cross_dataset}
\end{table*}

\begin{table}
\centering
\resizebox{\columnwidth}{!}{
\begin{tabular}{@{}llllllll@{}}
\toprule
 Method & Backbone   & Database & PQ    & mAP   & mIoU  \\ \midrule
 CLIP         & CLIP-ViT-base   & N/A      & 0.090  & 0.055 & 0.123 \\
 Retrieval    & CLIP-ViT-base   & ADE20k      & 0.117 & 0.071 & 0.150  \\
 \methodname      & CLIP-ViT-base   & ADE20k     & \textbf{0.127} & \textbf{0.075} & \textbf{0.173} \\ \midrule
 CLIP         & CLIP-ViT-large  & N/A    & 0.109 & 0.069 & 0.138 \\
 Retrieval    & CLIP-ViT-large  & ADE20k     & 0.158 & 0.098 & 0.215 \\
 \methodname      & CLIP-ViT-large  & ADE20k    & \textbf{0.161} & \textbf{0.103} & \textbf{0.222} \\
\bottomrule
\end{tabular}
}
\caption{Open vocabulary performance on ADE20k, training free, using Grounding DINO + SAM as mask proposal}
\label{tab:training_free}
\end{table}

\textbf{Setup:} The cross-dataset setup is fine-tuned on COCO panoptic annotations. The training-free setup does not use any panoptic segmentation annotations.  For constructing the retrieval feature database, we use the ADE20k~\cite{zhou2019semantic} train set and Google Open Image dataset~\cite{OpenImages} in separate settings. The evaluations are reported on the ADE20k validation set. Out of 150 classes in the ADE20k validation set, 70 are present in COCO. These classes serve as in-vocabulary classes and the rest of the classes are out-of-vocabulary. We experiment with different CLIP backbones such as CLIP-ViT-base, CLIP-ViT-large, CLIP-RN50x64, CLIP-ConvNeXt-large. We use Grounding-DINO-base for object detection and SAM-ViT-base for segmentation. Mask proposal generator in the cross-dataset setup benefits from fine-tuning on COCO. For the training-free setup, we experiment with different mask proposal methods such as ground truth mask, point prompt grid sampling with SAM, and Grounding DINO with SAM for ablation purposes.

\textbf{Baseline and Metrics:} We use FC-CLIP baseline in the cross-dataset setup and CLIP baseline for the training-free setup. For hyper-parameters in the FC-CLIP baseline, we the the same configuration used by \cite{yu2024convolutions}, setting $\alpha = 0.4, \beta = 0.8$. We use panoptic quality (PQ), mean intersection over union (mIoU), and mean average precision (mAP) as evaluation metrics.

\textbf{Results:} The performance of \methodname in cross-dataset setup is shown in Table~\ref{tab:cross_dataset}. When we use the ADE20k training set as a feature database, \methodname achieves an absolute improvement of +4.3 PQ and +4.5 PQ for CLIP-RN50x64 and CLIP-ConvNeXt-large backbone respectively. To check the robustness of \methodname with respect to domain shift in the feature database, we also run experiments using the Google Open image dataset as a feature database and evaluate on ADE20k. In this case, the absolute improvement over the FC-CLIP baseline is +0.9 PQ and +1.9 PQ for CLIP-RN50x64 and CLIP-ConvNeXt-large backbone respectively. The result demonstrates that \methodname helps improve performance even if the feature database is constructed using a different dataset compared to the test data. 

\methodname also improves performance in training-free setup as shown in Table~\ref{tab:training_free}. The results in Table~\ref{tab:training_free} are produced using ADE20k as a feature database and Grounding DINO+SAM as a mask proposal generator. \methodname achieves an absolute improvement of +3.7 PQ and +5.2 PQ for CLIP-ViT-base and CLIP-ViT-large backbone respectively. We also perform ablation of \methodname by removing CLIP-based classification in the training-free setup. The resulting system only relies on retrieval-based classification of out-of-vocabulary classes. Interestingly, we find that retrieval-based classification alone outperforms CLIP-only baseline for out-of-vocabulary classes, with +2.7 PQ and +4.9 PQ for CLIP-ViT-base and CLIP-ViT-large backbone respectively. This result demonstrates that retrieval itself can be a strong baseline because it is robust to domain shift between natural image CLIP features and masked image CLIP features.

\begin{table}
\centering
\resizebox{\columnwidth}{!}{
\begin{tabular}{@{}llllllll@{}}
\toprule
Mask Proposal &  Backbone   & PQ    & mAP   & mIoU  \\ \midrule
Ground Truth  & CLIP-ViT-base   & 0.211 & 0.133  & 0.276 \\ 
Grid Sampling + SAM  & CLIP-ViT-base   & 0.052 & 0.034 & 0.069 \\ 
Grounding DINO + SAM  & CLIP-ViT-base  & 0.127 & 0.075 & 0.173 \\  \midrule
Ground Truth  & CLIP-ViT-large  & 0.284 & 0.173 & 0.394 \\ 
Grid Sampling + SAM   & CLIP-ViT-large & 0.078 & 0.042 & 0.112 \\
Grounding DINO + SAM  & CLIP-ViT-large & 0.161 & 0.103 & 0.222 \\
\bottomrule
\end{tabular}
}
\caption{Impact of mask proposal quality of \methodname on training free setup, using ADE20k as feature database.}
\label{tab:mask_prop}
\end{table}

We demonstrate the impact of the mask proposal quality in the training-free setup in Table~\ref{tab:mask_prop}. The system achieves a PQ of 28.4 even with a ground truth mask with a CLIP-ViT-large backbone. Automatic mask generation with SAM performs poorly with a PQ of 7.8. The reason is that SAM is trained for interactive input with humans in the loop. Without human input, SAM masks are not class-aware. SAM may break up a single object into multiple fine masks. We mitigate this issue in the training-free setup by using open vocabulary object detection to construct class-aware bounding boxes and feeding them to SAM. This approach improves PQ to 16.1 in the training-free setup. 

The hyper-parameter tuning for ensemble coefficients is shown in Table~\ref{tab:hyper}. We find best performance with $\alpha = 0.4, \beta = 0.7, \gamma = 0.3$. \footnote{Case studies are available at \href{https://sigport.org/sites/default/files/docs/ICIP_OVPS_Supplementary.pdf}{this link}}

\begin{table}
\centering
\resizebox{0.8\columnwidth}{!}{
\begin{tabular}{cccc|cccc}
\toprule
$\alpha$ & $\beta$ & $\gamma$ & PQ   & $\alpha$ & $\beta$ & $\gamma$ & PQ    \\ \midrule
1.0       & 1.0      & 0.3      & 0.248  & 0.4       & 0.7      & 0.5      & 0.278 \\
0.5       & 0.7      & 0.3      & 0.303  & 0.4       & 0.7      & 0.4      & 0.297 \\
0.4       & 0.9      & 0.3      & 0.299  & 0.4       & 0.7      & 0.3      & \textbf{0.309}  \\
0.4       & 0.8      & 0.3      & 0.303  & 0.4       & 0.7      & 0.2      & \textbf{0.309}  \\
0.4       & 0.7      & 1.0      & 0.254  & 0.4       & 0.7      & 0.1      & 0.299  \\
0.4       & 0.7      & 0.7      & 0.278  & 0.4       & 0.7      & 0.0      & 0.264 \\
0.4       & 0.7      & 0.6      & 0.288  & 0.3       & 0.7      & 0.3      & 0.305  \\
\bottomrule
\end{tabular}
}
\caption{Hyper-parameter tuning, cross dataset setup with ADE20k as feature database}
\label{tab:hyper}
\end{table}

\section{Related Work}
Fully supervised methods typically involve training or fine-tuning the system on a dataset with pixel-level annotations. Ding et al.~\cite{ding2022open} use a trainable relative mask attention module to produce robust masked segment features from a frozen CLIP backbone. Xu et al.~\cite{xu2023open} propose combining the internal representation of pretrained text-to-image diffusion models and discriminative image-text models for open vocabulary panoptic segmentation. Liang et al.~\cite{liang2023open} fine-tune a CLIP backbone to improve alignment between text representation and masked image representation. Xu et al.~\cite{xu2023masqclip} use student-teacher self-training to improve mask generation for unseen classes and fine-tune CLIP to improve query feature representation. Yu et al.~\cite{yu2024convolutions} use a frozen CNN-based CLIP backbone for both mask proposal generation as well as classification. 

Weakly supervised methods are trained on image-level annotations. Luo et al.~\cite{luo2023segclippatchaggregationlearnable} train the system on image-text pairs using a semantic group module to aggregate patches with learnable image regions. He et al.~\cite{He_2023_CVPR} use self-supervised pixel representation learning guided by CLIP image-text alignment for semantic segmentation. Mukhoti et al.~\cite{Mukhoti_2023_CVPR} propose patch-level contrastive learning that learns alignment between visual patch tokens and text tokens. This approach generalizes to the open vocabulary setting without any training on pixel-level annotations. 

Training-free methods typically exploit pretrained models (e.g. CLIP) for open vocabulary segmentation without any fine-tuning on pixel-level or image-level annotations. Shin et al.~\cite{shin2022reco} construct a database of reference image segments using CLIP. During inference, the reference images are used for segmenting relevant segments from the input image. Karazija et al.~\cite{karazija2024diffusionmodelsopenvocabularysegmentation} generate synthetic reference images using a text-to-image diffusion model and perform segmentation by comparing input images with synthetic references. Wysoczanska et al.~\cite{Wysoczanska_2024_WACV} encodes small image patches separately to the vision encoder and computes class-specific similarity for an arbitrary number of classes. Then they perform patch aggregation, up-sampling, and foreground-background segmentation to produce segmentation for unseen classes. Gui et al.~\cite{gui2024knnclipretrievalenablestrainingfree} construct a feature database of masked segment features and use retrieval to perform panoptic segmentation on unseen categories. There are two key differences between their approach and our proposed method. Firstly, \cite{gui2024knnclipretrievalenablestrainingfree} uses one visual encoder for mask proposal generation and masked segment classification and a separate visual encoder to construct retrieval key features. We demonstrate that a single CLIP backbone with mask pooling can be used for all three tasks: mask proposal generation, retrieval key generation, and masked segment classification. Secondly, \cite{gui2024knnclipretrievalenablestrainingfree} rely on ground truth masks for constructing the feature database so their proposed approach cannot be extended to a new dataset where pixel-level annotation is unavailable. We use open vocabulary object detection combined with SAM for constructing the feature database and demonstrate that the improvement of our approach is robust to domain shift across multiple datasets.

\section{Conclusions}
In this work, we propose a retrieval-based method for improving open vocabulary panoptic segmentation. We construct a visual feature database using paired image-text data. During inference, we use masked segment features from the input image as query keys to retrieve similar features and associated class labels from the database. Classification scores for the masked segment are assigned based on the similarity between query features and retrieved features. The retrieval-based classification scores are combined with CLIP-based scores to produce the final prediction. The proposed approach improves PQ from 26.4 to 30.9 on ADE20k when fine-tuned on COCO. Even though the proposed method achieves reasonable performance in an open vocabulary setting, it remains vulnerable to the quality of mask proposal generation. Future work may focus on improving the quality of mask proposal generation for unknown classes.

\bibliographystyle{IEEEbib}
\bibliography{strings,refs}

\end{document}